\documentclass{article}

\PassOptionsToPackage{numbers,sort&compress}{natbib}
\usepackage[final,creativeai]{neurips_2026}

\usepackage[utf8]{inputenc} 
\usepackage[T1]{fontenc}    
\usepackage{booktabs}       
\usepackage{amsmath}
\usepackage{amsfonts}       
\usepackage{graphicx}
\usepackage{subcaption}
\usepackage{xcolor}         
\usepackage{microtype}      
\usepackage{hyperref}       

\newcommand{\compacttable}{%
  \renewcommand{\arraystretch}{0.94}%
  \setlength{\aboverulesep}{0.15ex}%
  \setlength{\belowrulesep}{0.25ex}%
}

\providecommand{\pms}[1]{{\,\scriptsize\textcolor{black!45}{$\pm$#1}}}

\newcommand{\blfootnote}[1]{%
  \begingroup
  \renewcommand{\thefootnote}{}%
  \def\theHfootnote{blfootnote}%
  \footnote{#1}%
  \addtocounter{footnote}{-1}%
  \endgroup
}

\title{FlockWorld: Toward Population-Scale Multi-Agent Video World Models through Artificial Life Simulation}

\author{
    Meryem Koksal \\
  Grinnell College\\
  Grinnell, United States \\
  \texttt{koksalme@grinnell.edu} \\
   \And
   Yunzhe Wang \\
    University of Southern California \\
   Los Angeles, United States \\
   \texttt{yunzhewa@usc.edu} \\
   \AND
   Volkan Ustun \\
   USC Institute for Creative Technologies \\
   Los Angeles, United States \\
   \texttt{ustun@ict.usc.edu} \\
}

\begin{document}

\maketitle
\begin{abstract}
Most action-conditioned video world models remain limited to a single player, while existing multiplayer models support only small, fixed groups of agents. 
We introduce FlockWorld, an artificial-life environment built on Reynolds's Boids algorithm for studying population-scale multi-agent video world models.
Specifically, FlockWorld renders synchronized egocentric crops, assigns distinct colors to preserve camera-agent identity across views, exposes rule-derived accelerations as actions, and lets the number of camera agents be configured independently of the size of the surrounding population.
We further introduce FlockDiT, an action-conditioned latent diffusion transformer trained with flow matching on FlockWorld data to jointly generate synchronized egocentric videos for multiple agents.
Our results suggest that joint generation benefits agent detection at density-matched settings and that diffusion forcing improves rollout fidelity. Together, FlockWorld and FlockDiT provide a controlled testbed for multi-agent video generation.%
\blfootnote{Codebase, configs, and rollout videos are available at \url{https://github.com/wangyz1999/flockworld}}

\end{abstract}


\section{Introduction}
Interactive video world models aim to learn controllable, persistent, and real-time simulators through action-conditioned video generation, with broad applications in interactive environment generation~\cite{bruce2024genie}, real-time neural game engines~\cite{alonso2024diffusion,wang2026matrix,valevski2025diffusion}, robotic control and planning~\cite{yang2023learning,ye2026world}, autonomous-driving simulation~\cite{russell2025gaia,agarwal2025cosmos}, and generative computer interfaces~\cite{zhuge2026neural}. 
Recent work such as Solaris~\cite{savva2026solaris}, Gamma-World~\cite{liu2026gamma}, and MIRA~\cite{hu2026multiplayer} has extended this paradigm to multiplayer settings, where generative world models must jointly represent the actions and observations of multiple agents within a shared environment (Appendix~\ref{app:related}).
However, scaling current multi-agent world models to massive populations and socially aware interactive environments with complex behavior remains underexplored. 
Achieving this capability could enable persistent virtual societies, population-scale training environments for embodied and social agents, and counterfactual simulations of complex collective systems. 

To investigate these challenges in a tractable setting, we turn to artificial-life simulation, in which agents act from local observations, influence one another within a shared environment, and produce population-level behavior through repeated interaction. In a flock, each agent's movement depends on those around it, giving rise to collective behavior. This offers a perspective on agency in which individual behavior and collective dynamics are inseparable, as each agent follows its own trajectory yet remains part of a larger whole. Recent environment-grounded LLM-agent simulations instantiate this structure in richer social settings, showing how shared worlds mediate individual and collective behavior~\cite{park2023generative,al2024project,wang2025implicit,ren2025simworld}.
In this paper, we study this structure in a controlled visual setting by introducing FlockWorld, a visually grounded multi-agent artificial-life environment for video world modeling built on Reynolds's Boids algorithm~\cite{reynolds1987flocks}, in which local inter-agent dynamics are directly reflected in agents' spatial configurations and trajectories.

Rule-based and emergent systems are a longstanding subject of generative art, in which the artist specifies a process and the work is produced by its execution~\cite{galanter2003}. Flocking is one such system, introduced by Reynolds as a distributed behavioral model. Creative agency is distributed by construction, since a vision radius and three rule weights are specified but no one determines the flock's shape at any given moment. FlockWorld retains these rules, following the parameterization of an open-source implementation~\cite{boidscode}, and adds a viewpoint, so the same flock can be observed from the perspectives of 2 to 30 of its own members, which we call camera agents, rather than from above. FlockDiT, the world model we introduce below, distributes agency further by giving each camera agent its own action stream, which modulates only that agent's view. Our two main failure modes are failures of that distribution. Rolling out each view independently invents camera agents the episode never contained, and one of our two conditioning schemes produces ten views that behave as a single scene cut into ten panels rather than ten separate observers.

FlockWorld supports a configurable population size and a camera-agent count that can be set independently of it. It provides synchronized egocentric observations and acceleration actions for each camera agent (Sec.~\ref{sec:flockworld}). We fix the population at 100 boids and evaluate configurations with 2 to 30 camera agents. We also introduce FlockDiT, an action-conditioned latent diffusion transformer that jointly generates multiple agents' views using temporally causal cross-agent attention (Sec.~\ref{sec:flockdit}). We study different schemes for representing agent identity and conditioning on actions, as well as the effects of diffusion forcing, single-agent pretraining, and a position-encoding background.

We first evaluate whether joint multi-agent generation improves upon generating each view independently (Sec.~\ref{sec:evaluation}). The joint model achieves higher detection rates than the independent single-agent baseline, with the clearest results at 10 and 30 camera agents, where detection density is controlled. We then examine the effects of different training and conditioning strategies (Sec.~\ref{sec:results}). Among the configurations tested, diffusion forcing most consistently improves rollout fidelity, reducing over-generation and improving cross-view agreement.

\section{FlockWorld}
\label{sec:flockworld}

\begin{figure}[t]
    \centering
    \includegraphics[width=0.85\linewidth]{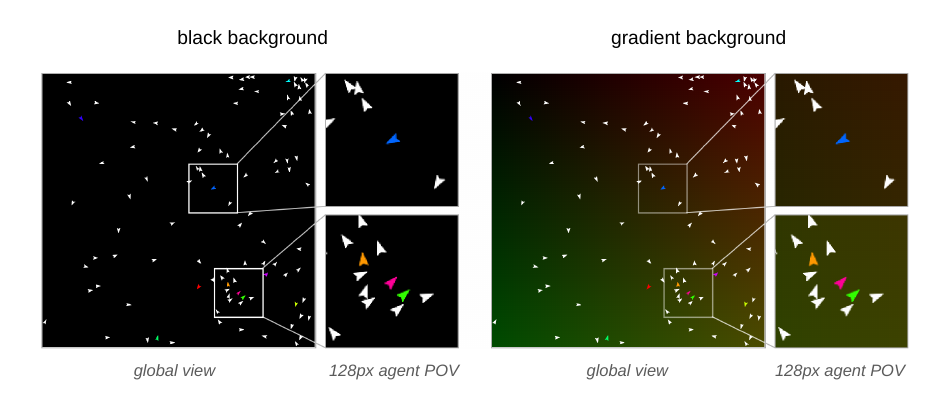}
    \caption{FlockWorld rendering conditions. A black background (left) provides no global-position cue, while a spatial gradient (right) encodes world position. Each condition shows the global view and two of its $128\times128$ agent-centered crops (boxed). Colored boids are identifiable camera agents, while white boids are ambient entities.}
    \label{fig:env_figure}
\end{figure}

We implement FlockWorld in JAX as an interactive 2D simulation of 100 boids in a $720\times720$ pixel arena, as illustrated in Fig.~\ref{fig:env_figure}. Boid motion follows three local Reynolds flocking rules~\cite{reynolds1987flocks}. \textbf{Alignment} encourages neighboring boids to move in similar directions, \textbf{cohesion} draws each boid toward the local flock center, and \textbf{separation} prevents collisions and overcrowding by repelling nearby boids. Each boid responds only to neighbors within a 25-pixel radius. We use reflecting boundaries and randomly initialize each boid's position, heading, and speed, discarding the first 60 steps as warm-up before rendering at 30 frames per second. Steering constants and the five-point boid geometry follow~\cite{boidscode}. We expose each camera agent's rule-derived acceleration as its action signal.



\subsection{Multi-Ego Video Data Generation}

Unless otherwise noted, we designate $P=10$ boids as camera agents and record a synchronized, agent-centered $128\times128$ crop for each over a 30-second simulation, producing $P$ partial views of the same flock. We use two rendering conditions. In \textbf{colored agents}, camera agent $i\in\{0,\ldots,P-1\}$ has HSV hue $h=i/P$ at full saturation and value while ambient boids remain white. \textbf{Colored agents with a gradient background} adds a dim global-position cue, with red increasing from left to right and green increasing from top to bottom. Fig.~\ref{fig:env_figure} illustrates the two colored-agent conditions. In the agent-count experiments, we vary $P$ from 2 to 30.



\subsection{Acceleration as Action Data}
\label{sec:actions}

At each simulation step, we use each camera agent's acceleration $(a_x,a_y)$, the net steering signal produced by the alignment, cohesion, and separation rules, as its action. One simulation step is one video frame, so to reach the four-times-lower latent frame rate the world model uses (Sec.~\ref{sec:flockdit}) we retain the initial acceleration and causally average each subsequent group of four steps.

\section{FlockDiT}
\label{sec:flockdit}

\subsection{Action-Conditioned Video World Model}
\paragraph{Latent Video Representation.} We model FlockWorld in a compact latent space using a smaller-capacity version of the Wan 2.2 causal VAE~\cite{vae,wanvae}. We train a separate VAE from scratch for each visual condition and freeze it before training the world model. The VAE compresses $128\times128$ video by $16\times$ spatially and $4\times$ temporally into $8$-channel $8\times8$ latent frames. The compression is causal, so the first latent frame encodes one video frame and each subsequent frame four. Architecture and training hyperparameters are in Appendix~\ref{app:hparams}.

\begin{figure}[t]
    \centering
    \includegraphics[width=0.85\linewidth]{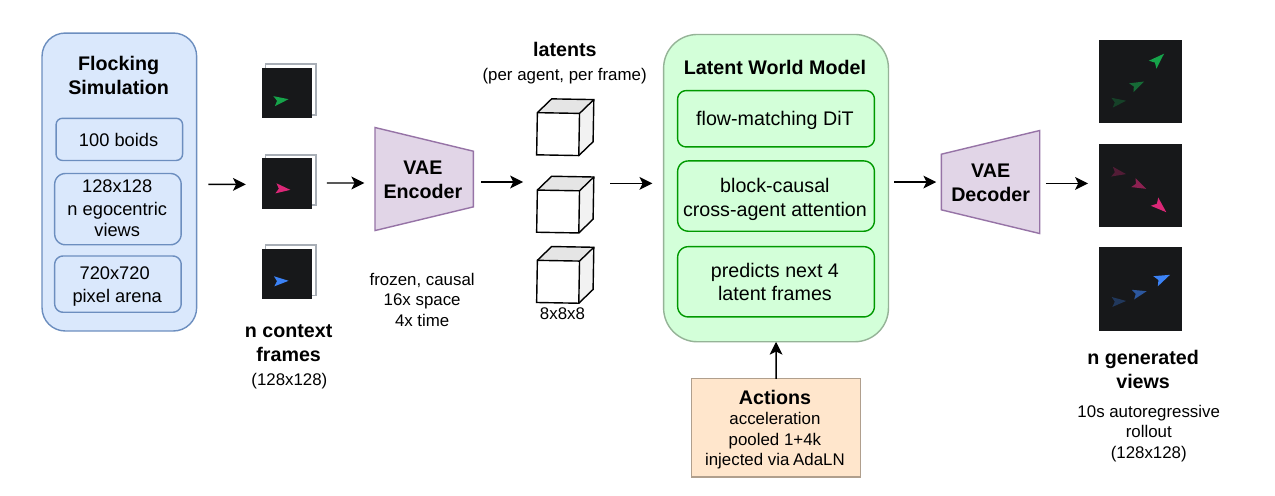}
    \caption{Overview of the FlockDiT model architecture.}
    \label{fig:arc_figure}
\end{figure}


\paragraph{Flow-Matching Transformer.} FlockDiT (Fig.~\ref{fig:arc_figure}) is a PyTorch adaptation of the Solaris single- and multi-player world model~\cite{savva2026solaris}. We retain its factorized 3D RoPE, AdaLN-zero conditioning, and block-causal attention, while removing image-to-video cross-attention for latent egocentric generation. Our diffusion transformer~\cite{attentiontransformer, peebles2023scalable} contains 8 blocks, has width 512, and comprises approximately 27.5M parameters. We use a block-causal temporal mask with full cross-agent attention, so a token at timestep $t$ may attend to all spatial tokens from every agent at timesteps $\tau\leq t$ but not at future timesteps. This preserves temporal causality while enabling information exchange across agent views. Actions reach the latent timeline by the pooling of Sec.~\ref{sec:actions}.

We train FlockDiT with the standard optimal-transport flow-matching objective~\cite{flowmatching}. Given a clean latent $z$, Gaussian noise $\varepsilon\sim\mathcal{N}(0,I)$, and $\sigma\sim\mathcal{U}[0,1]$, we define
\begin{equation}
x_\sigma=(1-\sigma)z+\sigma\varepsilon,\qquad
\mathcal{L}(\theta)=
\mathbb{E}\!\left[
\left\|v_\theta(x_\sigma,\sigma,a)-(\varepsilon-z)\right\|^2
\right],
\end{equation}
where $a$ denotes action conditioning.

\subsection{Multi-agent Conditioning}
FlockDiT jointly processes $P$ synchronized egocentric views (Appendix~\ref{app:setup}). We compare two coupled configurations that differ in how they represent agent identity and route actions.

\paragraph{Agent-embedded configuration.} All views use the same local spatial coordinates, and a learned agent embedding is added to each view's tokens. Each agent's acceleration modulates only its own tokens through AdaLN, making both agent identity and action assignment explicit.

\paragraph{Spatially tiled configuration.} Following MIRA~\cite{hu2026multiplayer}, we assign the latent token grids of the $P$ views to non-overlapping regions of a shared spatial RoPE grid~\cite{ropegridpositions}. For $P=10$, the views form a $2\times5$ layout. This tiles token coordinates rather than input pixels. Tile-specific RoPE positions provide implicit agent identity, so we disable the additive agent embeddings. To condition on actions, we add a learned agent tag to each encoded acceleration, mean-pool the tagged actions, and broadcast the resulting vector to all tokens through AdaLN. The model must therefore infer which component of the pooled action signal applies to each tile.

\subsection{Training Strategies}
\paragraph{Diffusion Forcing.}
By default we train with teacher forcing, in which the context frames remain clean and noise is applied only to the target frames. We additionally evaluate diffusion forcing~\cite{diffusionforcing}, in which every frame, including the context frames, is assigned an independent random noise level and contributes to the training loss. This exposes the model to corrupted context during training, matching the imperfect context it encounters during autoregressive rollout. Noise levels are drawn independently per frame and, in the \emph{embedded} configuration, per agent. In the \emph{tiled} configuration, all $P$ views of a frame are denoised jointly and therefore share one level~\cite{hu2026multiplayer}.

\paragraph{Two-Stage Training.} Modeling $P$ views jointly is a harder optimization problem than modeling one. We therefore pretrain the world model in a single-agent setting and initialize the multi-agent model from that checkpoint, then train on clips containing all $P$ agents, so the model learns the visual dynamics before adapting to multi-agent interactions~\cite{hu2026multiplayer}.

\section{Evaluation}
\label{sec:evaluation}

We evaluate each model on 28 held-out episodes. For each episode, we initialize the model with two ground-truth latent frames and autoregressively generate a 10-second rollout from the recorded actions, using a 50-step Euler solver per prediction window (Appendix~\ref{app:eval}). We decode generated latents to pixels and evaluate both \emph{per-view fidelity} to the recorded trajectory and \emph{cross-view consistency} among generated views.

\begin{figure}[t]
    \centering
    \begin{subfigure}{0.32\linewidth}
        \centering
        \includegraphics[width=\linewidth]{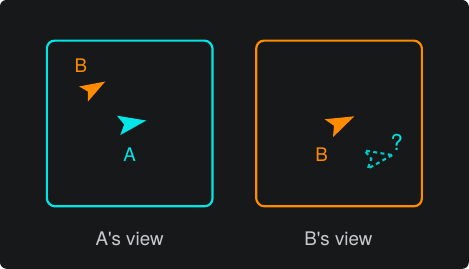}
        \caption{Reciprocity.}
        \label{fig:recip_figure}
    \end{subfigure}
    \hfill
    \begin{subfigure}{0.32\linewidth}
        \centering
        \includegraphics[width=\linewidth]{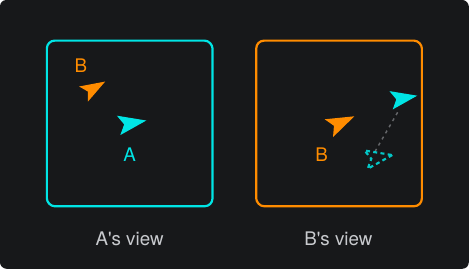}
        \caption{Displacement and motion error.}
        \label{fig:disperr_figure}
    \end{subfigure}
    \hfill
    \begin{subfigure}{0.32\linewidth}
        \centering
        \includegraphics[width=\linewidth]{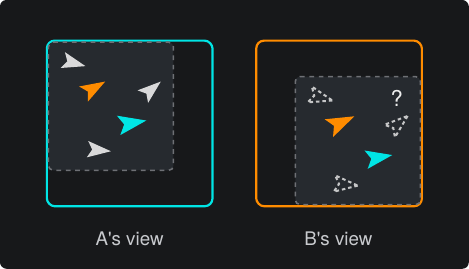}
        \caption{White-boid correspondence.}
        \label{fig:envmatch_figure}
    \end{subfigure}

    \caption{Cross-view consistency metrics (Appendix~\ref{app:eval}) for one agent pair seen
    from both views, showing (a)~mutual detection, (b)~disagreement about the pair's relative
    offset and its frame-to-frame change, and (c)~agreement about ambient white boids in the
    shared region.}
    \label{fig:consistency_metrics}
\end{figure}

\paragraph{Per-View Fidelity.} For each view, the \textbf{detection rate} is the fraction of ground-truth visible boids matched by a detection, the \textbf{position error} is the pixel distance between matched detections and their ground-truth positions, and \textbf{detections per frame} is the predicted number of visible boids, compared with the ground-truth count.

\paragraph{Cross-View Consistency.} We compare every pair of generated camera-agent views without access to the ground-truth state. As illustrated in Fig.~\ref{fig:consistency_metrics}, \textbf{reciprocity} measures whether two agents detect each other consistently, \textbf{displacement and motion error} measure the disagreement between their reciprocal estimates of relative position and frame-to-frame motion, and \textbf{white-boid correspondence} measures the spatial agreement between detections of ambient white boids in overlapping visible regions. We additionally report SSIM~\cite{ssim} over regions of overlapping views.

\paragraph{Evaluation Bounds.} The \textbf{floor} rolls out each view independently with the single-agent model, giving the agreement obtainable without information from the other views. The \textbf{ceiling} decodes ground-truth latents through the VAE and detects on the result, giving the best score the VAE and detector allow. Both use the same episodes, protocol, and detector as the models under test.

\section{Experimental Results}
\label{sec:results}

\subsection{Mechanism Comparison}
\label{sec:results-mechanisms}

\begin{table}[t]
  \centering
  \caption{Mechanism comparison at $P=10$ (black background). Metrics in $[0,1]$ are reported $\times100$, and $\pm$ values are 95\% confidence intervals across episodes (Appendix~\ref{app:eval}). \textbf{Bold} marks the best of Exps.\ 1-7 and any overlapping interval, and density and reciprocity are unranked. Exps.\ 3 and 5 average 27 and 26 episodes for Disp.\ err.\ and SSIM, and 27 and 25 for Motion err.}
  \label{tab:mechanisms}
  \small
  \compacttable
  \setlength{\tabcolsep}{2.5pt}
  \begin{tabular}{@{}l@{\hspace{2pt}}cccccccc@{}}
    \toprule
    & \shortstack{Det./\\frame}
    & \shortstack{Det.\ rate\\$\uparrow$}
    & \shortstack{Pos.\ err.\\px $\downarrow$}
    & \shortstack{Recip.\\{\ }}
    & \shortstack{White\\corr.\ $\uparrow$}
    & \shortstack{Disp.\ err.\\px $\downarrow$}
    & \shortstack{Motion err.\\px $\downarrow$}
    & \shortstack{SSIM\\$\uparrow$} \\
    \midrule
    Ceiling
      & 4.88 & 96.5\pms{0.4} & 1.496\pms{0.013}
      & 51.9\pms{7.0} & 85.6\pms{5.4} & 4.62\pms{1.77}
      & 1.05\pms{0.60} & 94.0\pms{0.7} \\
    \midrule
    1\ \ Baseline
      & 13.08 & \textbf{28.5}\pms{0.9} & 2.511\pms{0.048}
      & 17.4\pms{1.8} & 13.6\pms{1.2} & 62.02\pms{1.96}
      & 19.04\pms{2.05} & 43.1\pms{2.8} \\
    2\ \ Tiled
      & 23.14 & 22.6\pms{0.5} & 3.444\pms{0.033}
      & 43.5\pms{0.9} & 15.2\pms{0.5} & 64.08\pms{0.58}
      & 32.49\pms{0.66} & 37.1\pms{1.0} \\
    3\ \ DF
      & 4.51 & 26.7\pms{1.1} & 1.932\pms{0.039}
      & 22.6\pms{6.4} & \textbf{40.7}\pms{11.1}
      & \textbf{14.22}\pms{6.72} & \textbf{3.14}\pms{1.63}
      & \textbf{84.3}\pms{4.1} \\
    4\ \ 2-stage
      & 8.94 & \textbf{29.5}\pms{1.1} & 2.076\pms{0.070}
      & 46.4\pms{3.1} & 4.2\pms{1.1} & 61.06\pms{3.01}
      & 22.78\pms{2.23} & 63.2\pms{2.4} \\
    5\ \ Tiled+DF
      & 4.17 & 13.5\pms{1.1} & 2.594\pms{0.047}
      & 12.2\pms{3.9} & \textbf{33.5}\pms{12.0}
      & \textbf{15.48}\pms{10.60} & \textbf{1.14}\pms{0.44}
      & \textbf{85.8}\pms{3.8} \\
    6\ \ Tiled+DF+2-st.
      & 4.50 & 26.1\pms{1.0} & \textbf{1.827}\pms{0.034}
      & 11.4\pms{1.9} & 9.5\pms{3.8} & 61.22\pms{5.40}
      & 4.42\pms{0.53} & 79.2\pms{1.3} \\
    7\ \ Tiled+2-stage
      & 5.47 & 10.7\pms{0.8} & 2.845\pms{0.062}
      & 14.3\pms{2.7} & 7.9\pms{3.1} & 61.25\pms{5.90}
      & 7.84\pms{1.62} & 75.6\pms{4.3} \\
    \midrule
    Floor
      & 7.36 & 19.2\pms{1.1} & 2.506\pms{0.080}
      & 20.1\pms{2.2} & 5.3\pms{0.9} & 67.75\pms{2.07}
      & 12.79\pms{1.45} & 64.8\pms{3.7} \\
    \bottomrule
  \end{tabular}
\end{table}

The seven configurations are the embedded baseline (Exp.~1), tiling (2), diffusion forcing (3), two-stage training (4), and tiling with diffusion forcing (5), both (6), or two-stage training (7).

Detection density strongly confounds cross-view consistency metrics (Table~\ref{tab:mechanisms}). Experiment~2 achieves high reciprocity but renders 23.14 detections per frame against 5.04 in the ground truth, and its displacement error is close to the floor's, so raw reciprocity can reward over-rendering rather than agreement. Experiment~3, with diffusion forcing, matches the ground-truth density more closely and gives the strongest overall fidelity among the learned models at a usable detection rate, while Experiment~5 matches it on pixel and motion measures but detects half as many boids. Motion error, which tests the frame-to-frame change in relative offset and so removes static bias, cleanly separates the diffusion-forcing runs from the teacher-forced runs and the floor. Tiling alone worsens over-rendering, and adding two-stage training recovers its density calibration but not its detection rate. Overall, diffusion forcing is the most effective intervention tested.

\subsection{Qualitative Analysis of Rollout Behavior}
\label{sec:results-qualitative}

We inspect rollouts (Fig.~\ref{fig:filmstrip}, Appendix~\ref{app:qualitative}) for simulator-faithful behavior that aggregate metrics miss, namely focal-agent centering, color preservation, boundary dynamics, and whether the $P$ outputs are distinct views of the arena. Experiments 1, 3, 4, and 6 preserve centering, whereas 2, 5, 7, and the floor do not. Experiment 5 also loses agent color, explaining its low hue-based sighting count, and over-rendering appears as dense fields of spurious colored boids. In Experiments 5 and 7, boids often move continuously between adjacent tiles, making the $2\times5$ output resemble a partitioned panorama rather than ten independently conditioned viewpoints. Our budget-fixed reimplementation cannot establish that tiling causes this, but the failure shows that aggregate consistency metrics can reward coherent outputs built from an incorrect multi-view geometry.




\begin{table}[t]
  \centering
  \caption{Agent-count sweep, using the black background, the base
  configuration, and an equal 48-hour budget per run, with metrics in $[0,1]$ reported
  $\times100$. \textbf{Bold} marks density-matched settings where the model-floor gap exceeds
  the episode spread, and $^\dagger$ marks rows where the model renders
  ${\sim}2.5\times$ the floor's detections. The $P=10$ row is a different run from Exp.~1 of
  Table~\ref{tab:mechanisms} (Appendix~\ref{app:eval}).}
  \label{tab:agentcount}
  \small
  \compacttable
  \setlength{\tabcolsep}{4pt}
  \begin{tabular}{lccccccc}
    \toprule
    & \multicolumn{3}{c}{Detection rate $\uparrow$}
    & \multicolumn{3}{c}{Detections per frame}
    & Recip. \\
    \cmidrule(lr){2-4}\cmidrule(lr){5-7}
    $P$ & ceiling & model & floor & GT visible & model & floor & ceiling \\
    \midrule
    2
      & 97.0\pms{0.7} & 22.7\pms{2.5} & 20.1\pms{2.3}
      & 5.02 & 8.15 & 7.99 & 14.8 \\
    5
      & 96.6\pms{0.3} & 27.0\pms{0.9} & 19.7\pms{1.2}
      & 4.98 & 22.01$^\dagger$ & 8.06 & 47.2 \\
    8
      & 96.4\pms{0.4} & 31.8\pms{0.9} & 18.7\pms{1.2}
      & 5.04 & 18.45$^\dagger$ & 6.99 & 34.5 \\
    10
      & 96.5\pms{0.4} & \textbf{27.4}\pms{1.1} & 19.0\pms{0.9}
      & 5.04 & 7.11 & 7.48 & 51.9 \\
    20
      & 96.5\pms{0.3} & 28.1\pms{0.6} & 19.5\pms{0.7}
      & 5.11 & 18.33$^\dagger$ & 7.56 & 14.3 \\
    30
      & 96.5\pms{0.4} & \textbf{26.5}\pms{0.7} & 19.2\pms{0.8}
      & 5.10 & 7.68 & 7.39 & 7.9 \\
    \bottomrule
  \end{tabular}
\end{table}

\subsection{Number of Camera Agents}
\label{sec:results-agentcount}

To test whether the advantage of joint multi-view training persists as coordination scales, we sweep $P$ from 2 to 30 under a fixed wall-clock budget against the single-agent floor (Table~\ref{tab:agentcount}). Our model achieves a higher detection rate at every $P$, although results at $P=5$, $8$, and $20$ are confounded by its roughly $2.5\times$ higher detection density, and the margin at $P=2$ ($22.7$ against $20.1$) falls inside the episode-level spread. The strongest evidence therefore comes from the density-matched $P=10$ and $P=30$, where our model detects more agents at essentially the floor's detection density. It persists despite larger settings receiving less optimization (18 epochs at $P=30$ against 170 at $P=5$). We do not infer a scaling trend from reciprocity, since its ceiling varies non-monotonically with $P$, so each $P$ is compared only with its own references (Appendix~\ref{app:limitations}).

\begin{table}[t]
  \centering
  \caption{We evaluate the base configuration under both visual conditions, each against
  its own ceiling. Metrics in $[0,1]$ are reported $\times100$. The two conditions use
  separately trained frozen autoencoders, so their ceilings are not comparable.}
  \label{tab:conditions}
  \small
  \compacttable
  \setlength{\tabcolsep}{5pt}
  \begin{tabular}{lcccc}
    \toprule
    & \multicolumn{2}{c}{Black background}
    & \multicolumn{2}{c}{Gradient background} \\
    \cmidrule(lr){2-3}\cmidrule(lr){4-5}
    & ceiling & model & ceiling & model \\
    \midrule
    Detections / frame
      & 4.88 & 13.08 & 4.81 & 7.11 \\
    Detection rate $\uparrow$
      & 96.5\pms{0.4} & 28.5\pms{0.9}
      & 95.1\pms{0.6} & 21.3\pms{1.1} \\
    Position error px $\downarrow$
      & 1.496\pms{0.013} & 2.511\pms{0.048}
      & 1.504\pms{0.012} & 2.452\pms{0.065} \\
    Reciprocity $\uparrow$
      & 51.9\pms{7.0} & 17.4\pms{1.8}
      & 76.8\pms{5.2} & 17.4\pms{2.8} \\
    Displacement err.\ px $\downarrow$
      & 4.62\pms{1.77} & 62.02\pms{1.96}
      & 2.77\pms{0.25} & 56.08\pms{6.15} \\
    SSIM $\uparrow$
      & 94.0\pms{0.7} & 43.1\pms{2.8}
      & 94.9\pms{0.5} & 51.7\pms{5.1} \\
    \bottomrule
  \end{tabular}
\end{table}

\subsection{Position-encoding Background}
\label{sec:results-gradient}

We compare the gradient background with the black background using world models of the same architecture and budget, each against its own ceiling (Table~\ref{tab:conditions}). Because each condition needs its own autoencoder and ours differ in training length and data, the gradient condition's higher, more stable reciprocity ceiling cannot be attributed to its background alone, and neither absolute reciprocity nor the fraction of the ceiling recovered is comparable across conditions (Appendix~\ref{app:conditions}).

\section{Conclusion}

We introduced FlockWorld, an environment for population-scale multi-agent video world
modeling, and FlockDiT, which generates many agents' egocentric views jointly. Joint
generation reproduces more of the agents present than independent generation, an advantage
that persists as camera agents are added, and diffusion forcing is the most effective
training strategy we tested. Consistency metrics are confounded by rendered density and
should be judged against matched references. Scaling further will require sparser
cross-agent attention, such as over local neighborhoods.

\section{Acknowledgments}
Meryem Vera Koksal was supported by the National Science Foundation Research Experiences for Undergraduates (REU) program under Award Nos. 2150187 and 2446866. Any opinions, findings, and conclusions or recommendations expressed in this material are those of the authors and do not necessarily reflect the views of the National Science Foundation. The project or effort depicted was or is sponsored by the U.S. Army Combat Capabilities Development Command–Soldier Center under contract number W912CG-24-D-0001. The content of the information does not necessarily reflect the position or the policy of the Government, and no official endorsement should be inferred.


\bibliographystyle{plainnat}
\bibliography{ref}

\appendix
\newpage

\section{Related Work}
\label{app:related}

\paragraph{Interactive video world models.}
Action-conditioned video generation has emerged as a way to build interactive simulators. These models build on latent diffusion~\cite{rombach2022high} and diffusion transformers~\cite{peebles2023scalable} trained with flow matching~\cite{flowmatching}, and have been adapted into visual simulators beyond latent dynamics models for control~\cite{yang2023unisim,bar2025navigation,liang2024dreamitate}. Applications include video games, where models generate playable frames from controller input~\cite{bruce2024genie,parkerholder2024genie2,deepmind2025genie3,valevski2025diffusion,alonso2024diffusion,decart2024oasis,guo2025mineworld,zhang2025matrixgame,he2025matrixgame2,wang2026matrix,li2025hunyuangamecraft},
autonomous driving~\cite{hu2023gaia,russell2025gaia,agarwal2025cosmos}, robotic
control and planning~\cite{yang2023learning,li2025unified,gao2026dreamdojo,ye2026world},
and generative computer interfaces~\cite{zhuge2026neural}. Autoregressive rollout creates a train-test mismatch, since the model trains on clean context but generates from its own outputs. Diffusion forcing assigns each frame an independent
noise level~\cite{diffusionforcing}, while causal distillation and self-forcing supervise
the model on its own generations~\cite{yin2025causvid,huang2025self}. Such models could underpin generative game engines that respond to player actions~\cite{yu2025position}, but they condition on a single action stream, from one player or camera, and do not keep multiple simultaneous observers of one world consistent.

\paragraph{Multiplayer and multi-agent video world models.}
Modeling a world that several agents reshape at once is more recent. The World and
Human Action Model is trained on four-versus-four gameplay but conditions on one
player's controls, treating the others as environment~\cite{kanervisto2025world}, and
Multiverse simulates two-player racing by concatenating the players' frames
channel-wise~\cite{enigma2025multiverse}. Newer models condition on several agents at
once. Some model two players, or per-agent first-person views beyond two players, in
Minecraft~\cite{savva2026solaris,liu2026gamma}. Others handle multiple camera views of
cooperative play and robot manipulation~\cite{wu2026multiworld}, anchor DOOM levels to
editable 2D maps~\cite{po2026multigen}, bind actions to multiple subjects in 2D
games~\cite{pondaven2026actionparty}, scale multi-agent models from single-view
video~\cite{hu2026metaworld}, prompt each entity with language over shared
views~\cite{zhu2026incantation}, or tile per-player latents in representation
autoencoders~\cite{hu2026multiplayer}. These works show that joint multi-agent
generation is feasible, but at small, largely fixed agent counts and in environments
whose interactions are hard to vary or verify against ground truth. FlockWorld instead
sets the population size and the number of camera agents independently under exactly
known rules, so we can evaluate joint generation from $P=2$ to $P=30$ observers in a
100-agent population against a ceiling measured at the same $P$.

\paragraph{Multi-egocentric datasets and environments.}
Such models need synchronized first-person views of several agents, which mostly must be
collected. Real-world multi-egocentric datasets require people wearing cameras, as in
EgoHumans, with several Aria wearers during sports and
games~\cite{khirodkar2023egohumans}, and CASTLE 2024, with ten wearers over four days of
shared living~\cite{rossetto2025castle}. Multiplayer gameplay must likewise be recorded.
X-Ego-CS replays public Counter-Strike~2 matches to capture all ten players' views and
inputs~\cite{wang2025x}, GameplayQA manually aligns streamers' recordings of shared
matches for multi-video question answering~\cite{wang2026gameplayqa}, and MultiWorld
records hundreds of hours of human two-player play~\cite{wu2026multiworld}. Bots remove
the need for participants but require game-specific capture infrastructure, such as
Solaris's scripted-bot engine for two-player Minecraft~\cite{savva2026solaris}, a
pipeline modeled on it~\cite{liu2026gamma}, or 10,000 hours of bot-played Rocket
League~\cite{hu2026multiplayer}. In each case, the participants, the game, or the
pipeline sets the number of views. Simulators such as Melting
Pot~\cite{leibo2021meltingpot} and the CARLA-based OPV2V~\cite{xu2022opv2v} render
agent-centered observations synthetically, but for at most 16 agents and for policy
learning or cooperative perception, while Neural MMO supports larger populations with
tile-based rather than rendered observations~\cite{suarez2019neuralmmo}. FlockWorld is
fully synthetic. Its clips are simulated and rendered on the fly, so data is bounded by
compute rather than collection, and the camera-agent count is set independently of the
population size.

\section{FlockWorld Implementation Details}
\label{app:env}

This section documents the FlockWorld simulator in full. The environment is fully
synthetic and contains no human data.

\paragraph{Steering rules.}
The environment is a $720 \times 720$ pixel arena containing 100 boids that follow
the three steering rules of Reynolds~\cite{reynolds1987flocks}, namely alignment, cohesion, and
separation. These rules are computed over neighbors within a 25-pixel vision radius
and weighted by $1.1$, $1.0$, and $1.5$, respectively. Following the reference
implementation~\cite{boidscode}, the alignment rule additionally weights each
neighbor by $1.5^{\langle v_i,v_j\rangle}$, favoring neighbors heading in a similar
direction. When more than 32 candidate neighbors are available, each is retained
independently with probability $32/N$, where $N$ is the number of candidates, so
roughly 32 neighbors are expected to contribute. Each rule is implemented as a
Reynolds-style steering force, in which the desired velocity is set to the maximum speed and
the current velocity is subtracted. The resulting net steering force is clamped to a
maximum magnitude of $0.2$ per step, boid speeds are clamped to $[1,4]$ pixels per
step, and a velocity drag of $0.005$ per step is applied.

\paragraph{Boundaries and rendering.}
The arena has reflecting boundaries. When a boid reaches a boundary, its position is
clamped to 9 pixels from the canvas edge and the corresponding velocity component is
negated, causing the flock to bounce off the walls rather than wrap around. Each boid
is rendered as a 10-pixel oriented, notched triangle using the five-point geometry of
the reference implementation, with a 2-pixel white border marking the arena boundary.

\paragraph{Episodes and timing.}
At the beginning of each episode, positions are sampled uniformly within the arena,
inset by 7 pixels from the walls, headings uniformly from $[-\pi,\pi]$, and speeds
uniformly from the allowed range. Episodes are deterministically seeded, making each
recording reproducible. One simulation step corresponds to one video frame at 30
frames per second. We discard the first 60 warm-up steps so that recordings begin after the flock has settled into its typical dynamics. Our JAX simulator is a
reimplementation of the interactive flocking simulation from~\cite{boidscode},
matching its steering rules, parameter defaults, and boid geometry.

\paragraph{Egocentric views.}
We designate the first $P$ boids as camera agents. They are a
subset of the 100 simulated boids and are subject to the same flocking dynamics as
the rest. Each camera agent observes a $128 \times 128$ egocentric crop centered on
itself, with zero-padding where the crop extends beyond the arena boundary. The crop
center is locked to integer pixel coordinates, so the focal agent sits at the center
of its own view without sub-pixel jitter. Each episode therefore produces $P$
synchronized partial views of the same underlying flock. Alongside the rendered frames, the simulator records the
full state of every boid, including positions, velocities, accelerations, and
headings at every frame.

\paragraph{Training and evaluation data.}
Training data is generated on the fly because the simulator is fast. Each training
clip is a freshly simulated 21-frame ($0.7$ s) segment, corresponding to six latent
frames under the causal $1+4k$ temporal compression described in
Appendix~\ref{app:vae}. Clips are generated during data loading and encoded to
latents on the GPU. Each simulated episode is cut into several consecutive windows
(four during multi-agent training and sixteen during single-agent pretraining), so one
simulation supplies several temporally adjacent clips. This amortizes the cost of
simulating an episode, at the price of some correlation among the clips drawn from it.
However, because every episode is simulated fresh, the model never revisits a clip and
the effective training set is not fixed. The number of unique training clips is
limited primarily by the available training budget. For evaluation, we use a fixed
held-out set of recorded 30-second episodes, each containing 900 frames and $P$
synchronized views. The evaluation episodes are generated from seeds that are
disjoint from those used during training.

\section{Visual Conditions}
\label{app:conditions}

A central question in this work is whether the views of an episode can be made
mutually consistent, which requires the model to infer the relationships among the
camera agents. Boids in the base simulation are identical and white, so a crop of the
arena is anonymous. Aside from the arena boundary, it
carries no explicit information about its absolute location or about which agent is being observed. We therefore give each camera agent its own hue, spaced evenly around the
color wheel at intervals of $1/P$, while the non-camera boids stay white. Agent color
also enables our ground-truth-free consistency metrics (Appendix~\ref{app:eval}), which
identify camera agents in one another's views by hue. Note that the palette tightens as
$P$ grows, so hue identification becomes harder at large $P$ even for ground-truth
decodes.

Two visual conditions build on this and differ only in the background, which is either
black or a dim position-encoding gradient that supplies an absolute location cue. The mechanism comparison and the
agent-count sweep both use the black-background condition. The gradient-background
condition is evaluated as a single additional run. Because it requires its own
autoencoder and has no single-agent floor trained under matching rendering, it is
reported on its own rather than alongside the black-background results.

The two world models share an architecture and receive nearly the same optimization
budget, 393k against 394k steps, so the comparison of Sec.~\ref{sec:results-gradient} is
budget-matched at the level of the world model but not below it, since the black-background
autoencoder was trained for roughly 115k steps on a fixed recorded dataset, whereas the
gradient-background autoencoder was trained for roughly 887k steps on streamed clips
(Appendix~\ref{app:vae}). Because every ceiling is a decode through the condition's own
autoencoder passed through the hue-based detector, autoencoder quality sets the ceiling
directly, and the gradient condition's higher and more stable reciprocity ceiling cannot
be attributed to its background representation in isolation. This is why we compare each
model only with the ceiling measured under its own rendering, and why we do not read
across conditions from either the absolute reciprocity values or the fraction of the
ceiling a model recovers. Within its own condition, the gradient-background model renders
$7.11$ detections per frame against its ceiling's $4.81$ and sees more camera agents than
the black-background model, yet stays far below its ceiling in reciprocity, because the additional
detections it renders rarely form mutual cross-view matches.

\section{Problem Setup}
\label{app:setup}

We consider an environment observed by $P$ camera agents. The mechanism
comparison of Appendix~\ref{app:protocol} holds it at $P=10$ so that the conditioning
scheme, objective, and curriculum are the only things that vary, and we vary $P$
separately to measure how the model scales with the number of agents.
Each agent $p$ observes its own egocentric video stream of the shared arena
and receives an action stream consisting of its per-frame acceleration.
A frozen autoencoder compresses each video stream
spatially and temporally into a sequence of latent frames at one quarter of the
original video frame rate (Appendix~\ref{app:vae}). We index latent frames by $t$,
writing $z_t^p$ for agent $p$'s latent representation at step $t$ and $a_t^p$ for
its pooled action over the same step. The world model defines a distribution over
the next latent frames of all agents conditioned on their past latent frames and
actions,
\begin{equation}
\label{eq:world-model}
p_\theta\left( z_{t+1}^{1:P} \mid z_{\leq t}^{1:P}, a_{\leq t}^{1:P} \right).
\end{equation}
The model is rolled out autoregressively and the predicted latents are decoded back
into video frames. Training uses only rendered pixels and actions, and the model does
not receive the simulator state.

\section{Video Autoencoder}
\label{app:vae}

\paragraph{Architecture and compression.}
The autoencoder follows the causal 3D architecture of Wan
2.2~\cite{wanvae}. The input video is first patchified spatially using $2 \times 2$
patches and then passed through an encoder with three spatial downsampling stages,
two of which also downsample time. This produces a total compression of $16\times$
spatially and $4\times$ temporally. A $128 \times 128$ video containing $T$ frames is
mapped to a latent tensor of shape $8 \times T_{\mathrm{lat}} \times 8 \times 8$,
where
\begin{equation}
\label{eq:tlat}
T_{\mathrm{lat}} = 1 + \frac{T-1}{4}.
\end{equation}
Thus, the latent representation has 8 channels on an $8 \times 8$ spatial grid. The
first latent frame corresponds to one video frame, while each subsequent latent
frame corresponds to four video frames. All temporal convolutions are causal, so a
latent frame never depends on future video frames.

\paragraph{Action pooling.}
The same causal $1+4k$ structure determines how the per-frame actions are pooled
onto the latent timeline. Writing $a_i$ for the action at video frame $i$ and
$\tilde{a}_t$ for the action assigned to latent frame $t$,
\begin{equation}
\label{eq:action-pool}
\tilde{a}_t =
\begin{cases}
a_1, & t = 1, \\[2pt]
\dfrac{1}{4} \displaystyle\sum_{j=0}^{3} a_{\,4t-6+j}, & t \geq 2 ,
\end{cases}
\end{equation}
so the second latent frame receives the mean of video frames $2$ through $5$, the
third the mean of frames $6$ through $9$, and so on. The grouping is causal, so no
latent frame receives actions from video frames occurring after those it represents.

\paragraph{Training.}
We train two autoencoders (one per visual condition) from scratch at reduced capacity
rather than using released pretrained weights, since the encoder has to model its
condition's appearance. Both use an $L_1$ reconstruction loss and a KL penalty with
weight $10^{-6}$, and neither uses a perceptual or adversarial loss, because the
rendered environment is simple enough for $L_1$ reconstruction to recover the visual
details sharply. Both are trained on 17-frame clips, a different length from the
21-frame clips used later to train the world model, and both lengths satisfy the $1+4k$
requirement of the causal temporal compression, giving five and six latent frames
respectively. The black-background autoencoder reaches its best validation loss of $0.0042$ at epoch 81 of 100
(115{,}292 optimizer steps) on a fixed dataset of recorded episodes, while the
gradient-background autoencoder reaches $0.0046$ after roughly 887k steps on streamed
clips, so the two autoencoders differ in training length and data source even though they
share an architecture and optimizer. After
training each autoencoder is frozen, so the world model trains and rolls out entirely in
latent space, using the posterior mean as a deterministic latent encoding. Full VAE
hyperparameters are provided in Table~\ref{tab:vae-hparams}.

\section{FlockDiT Model Implementation Details}
\label{app:arch}

This section records how the velocity network $v_\theta$ converts latent frames into
tokens and how conditioning reaches them. The configuration itself is given in
Table~\ref{tab:wm-hparams}.

\paragraph{Tokenization.}
The model operates directly on the grid of latent vectors. A $1\times1\times1$
convolution maps each latent vector to the model width, making each latent vector a
single token. Each $8\times8$ latent frame therefore contributes $S=64$ tokens per
agent. For the mechanism comparison, with $P=10$ agents and $F=6$ latent frames, each
training clip contains
\begin{equation}
\label{eq:seq-len}
F\cdot P\cdot S = 6\cdot10\cdot64 = 3{,}840
\end{equation}
tokens. Tokens are arranged in frame-major order, so all agents' tokens for one latent
frame appear before the tokens of the next frame.

\paragraph{Positional encoding.}
We use factorized three-dimensional rotary positional embeddings
(RoPE)~\cite{ropegridpositions} over the temporal, height, and width axes. Each
attention head divides its dimensions among these three axes and applies the
corresponding rotations. Because RoPE encodes relative position, the same weights
can be used whether the $P$ views share one spatial coordinate frame or are arranged
as tiles in a larger grid (Appendix~\ref{app:multiagent}). Queries and keys are
RMS-normalized before applying the rotations.

\paragraph{Attention structure.}
The block-causal mask operates on blocks of $P\cdot S$ tokens, so the attention block
for each timestep spans all $P$ views at that timestep. This block width is what
makes the attention cost quadratic in the number of agents
(Appendix~\ref{app:limitations}). Each transformer block uses a standard pre-norm
architecture with a GELU feed-forward layer.

\paragraph{Action conditioning.}
The model is conditioned through adaptive layer normalization with zero
initialization (AdaLN-zero)~\cite{peebles2023scalable}. At each step, a conditioning
vector $c$ is formed by combining a sinusoidal embedding of the noise level $\sigma$
with an embedding of the actions, with each embedding passed through a small MLP.
For each transformer block, a linear projection maps $c$ to six modulation
parameters, namely a shift, scale, and gate for the attention sublayer and the same three
quantities for the feed-forward sublayer. These parameters are initialized so that
the modulation initially acts as the identity. The conditioning is defined per
latent frame and per agent and is broadcast over the 64 spatial tokens belonging to
that frame. Thus, each action can condition every spatial token of its corresponding
latent frame while remaining causal in time. The way the $P$ agents' actions are
combined depends on the multi-agent conditioning scheme described next.

\section{Multi-Agent Conditioning}
\label{app:multiagent}

This section records the implementation details of the two conditioning schemes
described in the main text. The embedded scheme follows Solaris~\cite{savva2026solaris}, using learned embeddings to distinguish agents whose views share the same spatial RoPE coordinates. The tiled scheme follows MIRA~\cite{hu2026multiplayer}, using a broadcast action vector with learned per-agent tags to associate actions with their corresponding view tiles. The tiled scheme also uses MIRA's warm-start curriculum.

\paragraph{Embedded.}
All agents share the same spatial RoPE coordinates, so the $P$ views, each an
$8\times8$ grid, occupy identical spatial positions. Their identities are encoded
by learned additive embeddings rather than by positional geometry.

\paragraph{Tiled.}
The broadcast conditioning vector is formed by embedding each agent's action, adding
a learned per-agent tag, and mean-pooling the results,
\begin{equation}
\label{eq:broadcast-action}
c_{\mathrm{act}} = \frac{1}{P} \sum_{p=1}^{P} g\left(e(a^p)+\omega_p\right),
\end{equation}
where $e$ is the action embedding, $\omega_p$ is the learned tag for agent $p$, and
$g$ is a shared projection. Since mean-pooling is order-invariant, the per-agent tag is
what preserves the association between an action stream and its view tile. Without
it, the pooled vector would carry no information about which action belongs to which
view.

\paragraph{Two-stage training.}
When the multi-agent model is initialized from the single-agent checkpoint, only the
multi-agent-specific parameters, namely the agent embeddings and the action-combination
head, are initialized from scratch, and the optimizer state is reset rather than
carried over from the pretraining stage.

\section{Experiments and Training Protocol}
\label{app:protocol}

Table~\ref{tab:experiments} summarizes the experimental runs. Experiment 1 is the
baseline, using embedded views, teacher forcing, and one-stage training. Experiments
2-4 each add one ingredient, either tiled views, diffusion forcing, or two-stage
training, so that each mechanism can be evaluated individually against the same
baseline. Experiments 2, 5, 6, and 7 form a $2\times2$ factorial design on tiled views, testing the effects of diffusion forcing and two-stage training. Experiment 2 uses neither technique, Experiment 5 uses diffusion forcing alone, Experiment 7 uses two-stage training alone, and Experiment 6 combines both. Experiment 6 therefore represents the full MIRA-inspired recipe~\cite{hu2026multiplayer}.

\begin{table}[t]
  \centering
  \caption{Experiment matrix. Every run trains the same world model (27.5M parameters
  embedded, 27.7M tiled)
  under the same 48-hour single-GPU allocation in the black-background condition. Only
  the listed ingredients differ. Two-stage runs divide the budget between
  single-agent pretraining and subsequent multi-agent training.}
  \label{tab:experiments}
  \small
  \compacttable
  \begin{tabular}{llllll}
    \toprule
    \# & Experiment & Views & Objective & Curriculum & Configuration \\
    \midrule
    1 & Baseline             & embedded & teacher forcing   & one-stage & \texttt{multi\_stream} \\
    2 & Tiled views          & tiled    & teacher forcing   & one-stage & \texttt{\ldots\_tiled} \\
    3 & Diffusion forcing    & embedded & diffusion forcing & one-stage & \texttt{\ldots\_df} \\
    4 & Two-stage            & embedded & teacher forcing   & two-stage & \texttt{\ldots\_twostage} \\
    5 & Tiled + DF           & tiled    & diffusion forcing & one-stage & \texttt{\ldots\_tiled\_df} \\
    6 & Tiled + DF + 2-stage & tiled    & diffusion forcing & two-stage & \texttt{\ldots\_triple} \\
    7 & Tiled + 2-stage      & tiled    & teacher forcing   & two-stage & \texttt{\ldots\_twostage\_tiled} \\
    \bottomrule
  \end{tabular}
\end{table}

\paragraph{Training protocol.}
All runs use streaming data (Appendix~\ref{app:env}), with clips simulated fresh
throughout training. Every run uses the same optimizer, learning rate, batch size,
and clip structure of two context and four target latent frames. The full
configuration is provided in Table~\ref{tab:wm-hparams}, and exact launch commands
are given in Appendix~\ref{app:repro}. Training is conducted under a fixed wall-clock budget rather than a fixed number of optimization steps. As a result, methods with higher per-step costs complete fewer optimization steps within the same training time. This allows us to compare methods based on the quality they achieve under the same time budget. All experiments are performed in the black-background
condition (Appendix~\ref{app:conditions}), which is required for the hue-based
consistency metrics.

\paragraph{Hyperparameter selection.}
We perform no hyperparameter search. The architectural and objective settings follow the Solaris implementation that we reimplement~\cite{savva2026solaris}, including factorized 3D RoPE with $\theta=10^4$, AdaLN-zero conditioning, block-causal attention, the rectified-flow objective, and the $[0,1000]$ timestep discretization. The embedded conditioning scheme follows Solaris~\cite{savva2026solaris}, while the tiled conditioning scheme follows MIRA~\cite{hu2026multiplayer}. From MIRA, we adopt the $2\times5$ view layout at $P=10$, the broadcast action vector with learned per-agent tags, the shared noise level across views of each frame, and the warm-start curriculum. The model size (8 blocks of width 512) and batch size (4) were chosen to fit a single-GPU run within the 48-hour allocation, rather than based on validation performance. The learning rate remains constant, with no warmup or decay.

Data-loading parameters were chosen to keep training efficient and do not affect the model, and all other settings are identical across experimental conditions. Differences between
conditions are therefore not attributable to per-run hyperparameter tuning. The
trade-off is that each condition is evaluated under a common fixed configuration
rather than at its individually optimized setting.

\section{Hyperparameters}
\label{app:hparams}

Tables~\ref{tab:wm-hparams} and~\ref{tab:vae-hparams} provide the complete
world-model and frozen-VAE configurations.

\begin{table}[t]
  \centering
  \caption{World model hyperparameters shared across all experiments.
  Table~\ref{tab:experiments} lists the per-experiment ingredient flags.}
  \label{tab:wm-hparams}
  \small
  \compacttable
  \begin{tabular}{ll}
    \toprule
    Hyperparameter & Value \\
    \midrule
    \multicolumn{2}{l}{\emph{Transformer architecture}} \\
    Parameters                & 27.5M embedded / 27.7M tiled \\
    Hidden dimension          & 512 \\
    Blocks                    & 8 \\
    Attention heads           & 8 (head dim.\ 64, QK RMS-norm) \\
    Feed-forward              & 2048, GELU \\
    Patch size                & 1 (one token per latent vector, 64 tokens/frame/agent) \\
    Positional encoding       & factorized 3D RoPE ($\theta=10^4$) \\
    Attention                 & block-causal in time, full within a timestep \\
    Conditioning              & AdaLN-zero (noise level $+$ action embedding) \\
    \midrule
    \multicolumn{2}{l}{\emph{Multi-agent}} \\
    Agents ($P$)              & 10 (mechanism comparison) and $2,5,8,10,20,30$ (scaling sweep) \\
    Action dimension          & 2 ($a_x,a_y$, raw and unnormalized) \\
    Embedded scheme           & own-action AdaLN $+$ additive agent embedding \\
    Tiled scheme              & RoPE tiling ($2\times5$ at $P=10$), broadcast actions, no agent embedding \\
    \midrule
    \multicolumn{2}{l}{\emph{Inputs}} \\
    Latent frame              & $8\times8$, 8 channels (from $128\times128$ RGB) \\
    Training clip             & 6 latent frames (2 context $+$ 4 target) $=$ 21 video frames \\
    Sequence length           & $F\!\cdot\!P\!\cdot\!S$ ($6\times10\times64=3{,}840$ at $P=10$) \\
    Latent normalization      & per-channel mean/std (8 calibration batches at startup) \\
    \midrule
    \multicolumn{2}{l}{\emph{Objective and sampling}} \\
    Objective                 & rectified flow, velocity prediction \\
    Noise levels              & uniform over 1001 levels, per frame (and per agent) \\
    Teacher forcing           & context clean, loss on target frames \\
    Diffusion forcing         & all frames noised independently, loss on all frames \\
    Sampling                  & Euler, 50 steps, context held clean, no CFG \\
    Rollout                   & sliding window, 4 new frames per step, 2-frame context \\
    \midrule
    \multicolumn{2}{l}{\emph{Optimization}} \\
    Optimizer                 & AdamW \\
    Learning rate             & $2\times10^{-4}$, constant (no warmup, no decay) \\
    Betas                     & $(0.9,0.999)$ \\
    Weight decay              & $10^{-6}$ \\
    Gradient clipping         & 1.0 (global norm) \\
    Mixed precision           & AMP \\
    Batch size                & 4 \\
    EMA                       & none \\
    \midrule
    \multicolumn{2}{l}{\emph{Streaming training}} \\
    Windows per epoch         & 20,000 (freshly simulated) \\
    Windows per episode       & 4 (16 in single-agent pretraining) \\
    Validation windows        & 64 (fixed) \\
    Budget                    & 48-h allocation, training capped at 2,860 min \\
                              & (two-stage runs, 710 min $+$ 2,150 min) \\
    Checkpointing             & every 1000 steps (keep 5) and every 2 epochs \\
    Seed                      & 42 \\
    \bottomrule
  \end{tabular}
\end{table}

\begin{table}[t]
  \centering
  \caption{Video autoencoder hyperparameters. The architecture, loss, and optimizer rows
  are shared by both condition-specific autoencoders, while the training-data and
  training-length rows describe the black-background checkpoint (epoch 81, validation loss
  0.0042), which every run except the gradient-background one uses. The
  gradient-background autoencoder was trained for longer and on streamed rather than
  recorded clips (Appendix~\ref{app:vae}).}
  \label{tab:vae-hparams}
  \small
  \compacttable
  \begin{tabular}{ll}
    \toprule
    Hyperparameter & Value \\
    \midrule
    Architecture                    & causal 3D convolutional VAE (Wan-style) \\
    Base width                      & 32 (channel multipliers 1,2,4,4 and 2 residual blocks/stage) \\
    Input patchify                  & $2\times2$ spatial \\
    Compression                     & $16\times$ spatial, $4\times$ temporal (causal, $1+4k$) \\
    Latent                          & 8 channels, $128^2\times T \rightarrow 8^2\times(1+(T-1)/4)$ \\
    Loss                            & $L_1$ reconstruction $+$ KL, weight $10^{-6}$ \\
    Perceptual / adversarial terms  & none \\
    Training clips                  & 17 frames, $128\times128$ (5 latent frames) \\
    Optimizer                       & AdamW, lr $10^{-4}$ constant (no schedule), \\
                                    & betas $(0.9,0.999)$, weight decay $10^{-6}$ \\
    Batch size                      & 32 \\
    Precision                       & bf16-mixed \\
    Gradient clipping               & 1.0 \\
    Training length                 & 100 epochs (best checkpoint at epoch 81, step 115{,}292) \\
    Training data                   & recorded black-background episodes, \\
                                    & 10\% validation split \\
    Inference                       & frozen, deterministic (posterior mean) \\
    \bottomrule
  \end{tabular}
\end{table}

\section{Evaluation Protocol and Metrics}
\label{app:eval}

\paragraph{Protocol.}
Each model is evaluated on 28 held-out episodes. Given the first two latent frames of
all $P$ views as context, the model generates a 10-second autoregressive rollout
using 50 Euler steps per prediction window while replaying the agents' recorded
actions. A 10-second rollout corresponds to 75 latent frames, which decode to 297
video frames under the causal $1+4k$ compression. The predicted latents are decoded
to pixels before metric computation. Because the model replays the actions from the recorded episode, a perfect prediction would reproduce the ground-truth trajectory. In practice, generated rollouts can diverge from the reference. Our metrics therefore distinguish errors relative to the ground truth, which are expected in generative models, from inconsistencies between views of the same scene, which indicate a failure of cross-view consistency.

\paragraph{Uncertainty and error bars.}
Each metric is computed separately for each of the 28 held-out episodes. We report
the mean across episodes with an error bar equal to a 95\% confidence interval
on that mean, computed with Student's $t$. The factor of variability it captures is
the choice of held-out episode, that is, which flock realization the model is asked to
continue.

\noindent Three qualifications apply.

\begin{itemize}
  \item \textbf{No variation across training runs.} Each condition corresponds to a
    single training run with a fixed seed, so the error bars capture nothing about
    initialization or data order. We report no variance across training seeds and
    perform no hypothesis tests. Differences that are small relative to the
    episode-level spread should not be read as ordering the mechanisms.
  \item \textbf{The interval is over episodes only.} It is estimated from 28 samples
    under a normality assumption, and indicates the precision of the episode mean
    rather than supporting a significance test across conditions.
  \item \textbf{Effective sample size varies by metric.} Cross-view metrics are
    defined only on frames where the required sightings or reciprocal observations
    occur, and episodes without such events are excluded from that metric's mean and
    interval. The effective sample size is therefore below 28 for models that rarely
    render other camera agents, which is also why we track volume counts alongside
    every rate.
\end{itemize}

\paragraph{Checkpoint selection.}
Where a scored epoch checkpoint is available we evaluate the one with the lowest
validation loss, as for the agent-count sweep at $P\geq5$. For the mechanism comparison,
the floor, the gradient-background run, and the $P=2$ point of the agent-count sweep,
the available checkpoint is the final
mid-epoch snapshot, which carries no validation
score because the validation pass runs only at epoch boundaries. Those runs are
therefore evaluated at the last state they reached before the wall-clock cutoff. Under
the fixed-budget protocol of Appendix~\ref{app:protocol} both cases compare runs near
the end of the same budget rather than at individually tuned stopping points. The two
kinds of checkpoint are nevertheless mixed within Table~\ref{tab:agentcount}, whose model
rows at $P\geq5$ are best-validation checkpoints while the floor it is compared against
is a final snapshot, so the comparison should be read with that asymmetry in mind.

The agent-count sweep is also a separate training campaign from the mechanism
comparison, so its $P=10$ point and Experiment 1 of Table~\ref{tab:experiments} are two
different runs of the same configuration rather than one run reported twice. They differ
mainly in rendered density, at $7.11$ against $13.08$ detections per frame, and that
difference is far larger than the run-to-run variation attributable to sampling. Rolling
the single-agent floor out twice from the same checkpoint on the same 28 episodes moves
its detection rate by $0.2$ points and its density by $0.12$ detections per frame.
Rendered density at a fixed configuration is therefore only partly reproducible across
training runs, and the density-matched comparisons of Table~\ref{tab:agentcount} should
be read with that caveat.

\paragraph{Notation.}
Let $\mathcal{A}$ denote the set of camera agents, with $|\mathcal{A}|=P$, and let
$t=1,\ldots,T$ index decoded video frames. Each crop is $S_{\mathrm{px}}=128$ pixels
square with center
\begin{equation}
\label{eq:crop-center}
c_0 = \left( \frac{S_{\mathrm{px}}}{2}, \frac{S_{\mathrm{px}}}{2} \right).
\end{equation}
All metrics operate on decoded pixels using the detector described in
Appendix~\ref{app:eval-details}. For view $a$ at frame $t$, the detector returns a
set of detections $\mathcal{D}_{a,t}$. Each detection has a sub-pixel centroid and is
assigned either a camera-agent identity or the label \emph{white} for anonymous
boids.

All metrics are computed separately within each episode and then averaged over the
28 held-out episodes. Thus, for ratio-valued metrics, we first compute the ratio for
each episode and then average the resulting values, rather than pooling all frames
and views across the evaluation set. This also determines the episode-level spread
reported by the error bars.

\subsection{Per-View Fidelity (versus Ground Truth)}

The first tier compares each generated view with the corresponding ground-truth
episode. Let $\mathcal{V}_{a,t}$ denote the boids that are visible in the
ground-truth projection into view $a$ at frame $t$. Each ground-truth position is
matched to its nearest detection,
\begin{equation}
\label{eq:nn-residual}
r_{a,t}(g) = \min_{q\in\mathcal{D}_{a,t}} \left\lVert q-g\right\rVert_2,
\qquad g\in\mathcal{V}_{a,t},
\end{equation}
with $r_{a,t}(g)=\infty$ if the view contains no detections. A match is considered
correct when $r_{a,t}(g)\leq\tau_{\mathrm{gt}}=6$ pixels. \textbf{Detection rate} is
the fraction of ground-truth-visible boids that are detected,
\begin{equation}
\label{eq:detection-rate}
\mathrm{DR} = \frac{
  \sum_{a,t} \left| \left\{ g\in\mathcal{V}_{a,t}:
    r_{a,t}(g)\leq\tau_{\mathrm{gt}} \right\} \right|
}{
  \sum_{a,t}|\mathcal{V}_{a,t}|
}.
\end{equation}
\textbf{Position error} is the mean residual over matched boids,
\begin{equation}
\label{eq:position-error}
\mathrm{PE} = \operatorname{mean} \left\{ r_{a,t}(g):
  g\in\mathcal{V}_{a,t}, \ r_{a,t}(g)\leq\tau_{\mathrm{gt}} \right\}.
\end{equation}
Because position error is computed only on matched detections, it must be
interpreted together with detection rate. A model that renders very few boids may
achieve a low position error simply because it is evaluated on a small, relatively
easy subset. 

\textbf{Detections per frame} measures over- and under-rendering by
comparing the mean number of detected boids with the mean number of
ground-truth-visible boids,
\begin{equation}
\label{eq:det-per-frame}
\frac{1}{PT}\sum_{a,t}|\mathcal{D}_{a,t}|
\qquad \text{versus} \qquad
\frac{1}{PT}\sum_{a,t}|\mathcal{V}_{a,t}|.
\end{equation}
These metrics can legitimately decrease when a model diverges from the reference
trajectory while remaining internally consistent. They therefore measure fidelity
rather than cross-view consistency.

\subsection{Cross-View Consistency (Ground-Truth-Free)}

The second tier does not use the simulator state. Instead, the generated views
evaluate one another. When agent $b$ is detected in agent $a$'s view, its hue
identifies it and its position relative to the crop center provides a spatial claim.
Excluding the focal agent, we define
\begin{equation}
\label{eq:offset}
d_{ab}(t) = p_{a\rightarrow b}(t)-c_0,
\end{equation}
where $p_{a\rightarrow b}(t)$ is the detected position of agent $b$ in view $a$. This
represents the claim that ``$b$ lies at offset $d_{ab}$ from $a$.'' Let
$s_{ab}(t)\in\{0,1\}$ indicate whether such a sighting exists. Because the crops are
centered on the agents and use the same $1{:}1$ scale, a consistent pair of views
should make mirrored claims, $d_{ba}(t)\approx-d_{ab}(t)$. We therefore define the
total number of sightings and reciprocal sightings over unordered agent pairs,
\begin{equation}
\label{eq:volumes}
N_{\mathrm{sight}} = \sum_{a<b}\sum_t \left( s_{ab}(t)+s_{ba}(t) \right),
\qquad
N_{\mathrm{recip}} = \sum_{a<b}\sum_t s_{ab}(t)s_{ba}(t).
\end{equation}
\textbf{Reciprocity} is the fraction of sightings that are mutual,
\begin{equation}
\label{eq:reciprocity}
\mathrm{Recip} = \frac{2N_{\mathrm{recip}}}{N_{\mathrm{sight}}}.
\end{equation}
The factor of two accounts for the two sightings contributed by a mutual pair and
makes the metric equal to one when every sighting is reciprocated.

Let $\mathcal{R}$ denote the set of reciprocal $(a,b,t)$ triples.
\textbf{Displacement error} measures disagreement in the estimated relative position
of the two agents,
\begin{equation}
\label{eq:displacement}
E_{\mathrm{disp}} = \frac{1}{|\mathcal{R}|}
\sum_{(a,b,t)\in\mathcal{R}} \left\lVert d_{ab}(t)+d_{ba}(t) \right\rVert_2.
\end{equation}
The error is zero exactly when the two views agree on the relative offset.
\textbf{Motion error} applies the same test to changes in the relative offset
between consecutive frames. It is evaluated only when the pair is reciprocal at both
$t-1$ and $t$, denoted by $\mathcal{R}^2$,
\begin{equation}
\label{eq:motion}
E_{\mathrm{motion}} = \frac{1}{|\mathcal{R}^2|}
\sum_{(a,b,t)\in\mathcal{R}^2}
\left\lVert \left[ d_{ab}(t)-d_{ab}(t-1) \right]
          + \left[ d_{ba}(t)-d_{ba}(t-1) \right] \right\rVert_2.
\end{equation}

A reciprocal pair also defines an approximate pixel-to-pixel mapping between the two
crops. We use the symmetrized relative offset
\begin{equation}
\label{eq:pose}
\hat d_{ab}(t) = \frac{1}{2} \left( d_{ab}(t)-d_{ba}(t) \right),
\end{equation}
so that a point $p_a$ in view $a$ maps approximately to $p_a-\hat d_{ab}$ in view
$b$. This mapping enables additional consistency checks.

\textbf{White correspondence} measures whether the two views agree about anonymous
white boids that fall within their shared region. Let
$\mathcal{W}_{a\rightarrow b}$ denote the white detections from view $a$ that fall
inside view $b$ under the inferred mapping, and define $\mathcal{W}_{b\rightarrow a}$
analogously. We match these detections by optimal assignment on
$\left\| (w_a-\hat d_{ab})-w_b \right\|_2$, accepting matches within
$\tau_{\mathrm{w}}=8$ pixels and rejecting ambiguous matches as described in
Appendix~\ref{app:eval-details}. Writing $m_{ab}(t)$ for the number of accepted
matches in a reciprocal triple, the correspondence score is
\begin{equation}
\label{eq:white-corr}
C_{\mathrm{white}} = \frac{
  \sum_{(a,b,t)\in\mathcal{R}} m_{ab}(t)
}{
  \sum_{(a,b,t)\in\mathcal{R}} \max \left( |\mathcal{W}_{a\rightarrow b}|,
                                 |\mathcal{W}_{b\rightarrow a}| \right)
}.
\end{equation}

\textbf{Pixel consistency} directly compares the overlapping image regions. We round
the symmetrized pose to an integer shift $(\delta_x,\delta_y) =
\operatorname{round}(\hat d_{ab})$, crop both decoded frames to the resulting overlap
region, and discard pairs with fewer than 8 pixels of overlap along either axis. On
the remaining region, we compute
\begin{equation}
\label{eq:psnr}
\mathrm{PSNR} = -10\log_{10}\operatorname{MSE},
\end{equation}
using pixel values in $[0,1]$, and mean-windowed SSIM~\cite{ssim} using
$7\times7$ box windows with $C_1=0.01^2$ and $C_2=0.03^2$, averaged over color
channels. Both quantities are recorded in the released evaluation outputs, and the tables in
this paper report SSIM, which is the more informative of the two on this content because
the frames are dominated by the black background.

\paragraph{Rates and volume counts.}
Cross-view rates depend on what the model generates, since a model that rarely renders other agents can produce fewer opportunities for cross-view consistency failures, making its rates look artificially favorable. We therefore track the corresponding volume counts alongside
the rates, including $N_{\mathrm{sight}}$, $N_{\mathrm{recip}}$, the denominator of
the white-correspondence metric, and the number of pixel-overlap events. These counts
determine each metric's effective sample size and are recorded in full in the released
evaluation outputs, while the tables in this paper report the rates and include counts only when they help interpret a comparison.

\paragraph{Ceiling and floor.}
We use two reference rows to help interpret the evaluation metrics. The \emph{ceiling}
decodes the ground-truth latents through the same frozen VAE and applies the identical detector and metrics. It therefore represents the best performance achievable within the latent representation and detection pipeline. It is below perfect performance because detector and hue-identification errors can reduce the measured scores. The \emph{floor} rolls out each view independently using a single-agent model, so the $P$ views are generated without any shared information. It therefore measures the level of cross-view agreement obtainable without joint modeling. Some agreement is still expected because all views start from the same ground-truth context and are generated under the same simulation conditions. Improvement over this floor indicates that jointly modeling the views improves cross-view consistency compared with generating each view independently. Both references use the same episodes, evaluation protocol, and detector as the models under test. The gradient-background run has a ceiling but no floor, since no single-agent model was trained under that rendering.

\section{Evaluation Details}
\label{app:eval-details}

\paragraph{Boid detector.}
Generated and reference frames are processed using the same detector. A pixel is
considered foreground when its brightness, $\max(R,G,B)$, exceeds $0.25$. Connected
components of foreground pixels form detections, whose positions are computed using
brightness-weighted centroids. A detection's identity is determined from the hue $h$
of its mean RGB value,
\begin{equation}
\label{eq:identity}
\hat{k}(h) = \operatorname{round}(hP)\bmod P.
\end{equation}
The identity is accepted only when the circular hue distance from $\hat{k}/P$ is at
most $\tau_{\mathrm{hue}}/P$, with $\tau_{\mathrm{hue}}=0.25$. Detections with
insufficient saturation ($\max-\min<0.05$ across channels) are classified as white
boids. We apply three filters to suppress artifacts from the arena boundary. Highly
elongated detections are removed as border lines, detections with large spatial
extent are removed as corners, and small detections touching the image boundary are
removed as border clippings. In cross-view metrics, detections within
$\tau_{\mathrm{focal}}=8$ pixels of the crop center are excluded because they
correspond to the view's focal agent.

Because generated colors can flicker across frames, detections are linked into
tracks using nearest-neighbor matching within 8 pixels. Each track is assigned a single identity based on votes across the full track, so brief hue changes do not create false sightings.

\paragraph{Matching.}
Ground-truth projections are matched to detections using nearest-neighbor matching
within 6 pixels. White-boid correspondence uses optimal assignment under the
inferred pixel mapping, with a maximum matching distance of 8 pixels. To avoid
ambiguous matches, a candidate match is discarded when its cost is not sufficiently
smaller than the second-best candidate, specifically when its cost exceeds $0.7$
times the second-best cost. All detector and matching parameters are summarized in
Table~\ref{tab:detector}.

\begin{table}[t]
  \centering
  \caption{Detector, matching, and identity parameters. All metrics in
  Appendix~\ref{app:eval} use these settings, and the ceiling and floor references
  use the same pipeline.}
  \label{tab:detector}
  \small
  \compacttable
  \begin{tabular}{lll}
    \toprule
    Parameter & Value & Role \\
    \midrule
    \multicolumn{3}{l}{\emph{Detection}} \\
    Foreground threshold        & $\max(R,G,B)>0.25$ & separates boids from background \\
    Minimum component size      & 2 px               & rejects single-pixel noise \\
    Maximum elongation          & 5                  & rejects arena border lines \\
    Maximum extent              & 40 px              & rejects arena corners \\
    Border band / minimum size  & 5 px / 25 px       & rejects border clippings \\
    Centroid                    & brightness-weighted & sub-pixel position \\
    \midrule
    \multicolumn{3}{l}{\emph{Identity}} \\
    Achromatic threshold        & $\max-\min<0.05$   & labels a detection white \\
    Hue palette                 & $h=k/P$, $k<P$     & camera-agent identity \\
    Identity tolerance          & $0.25/P$           & rejects off-palette hues \\
    Track link radius           & 8 px               & links detections across frames \\
    Identity vote window        & whole track        & robust to hue flicker \\
    Focal exclusion radius      & 8 px               & removes the view's own agent \\
    \midrule
    \multicolumn{3}{l}{\emph{Matching and pixel comparison}} \\
    Ground-truth match radius   & 6 px               & Eqs.~\eqref{eq:detection-rate}-\eqref{eq:position-error} \\
    White match radius          & 8 px               & Eq.~\eqref{eq:white-corr} \\
    Ambiguity margin            & 0.7                & excludes ambiguous white matches \\
    Minimum overlap per axis    & 8 px               & Eq.~\eqref{eq:psnr} \\
    PSNR data range             & 1.0                & pixels in $[0,1]$ \\
    SSIM window                 & $7\times7$ box     & $C_1=0.01^2$, $C_2=0.03^2$ \\
    \bottomrule
  \end{tabular}
\end{table}

\section{Qualitative Rollouts}
\label{app:qualitative}

\begin{figure}[t]
  \centering
  \setlength{\tabcolsep}{3pt}
  \begin{tabular}{@{}l@{\hspace{5pt}}c@{}}
    \rotatebox[origin=c]{90}{\small Ceiling} &
      \includegraphics[width=0.92\linewidth]{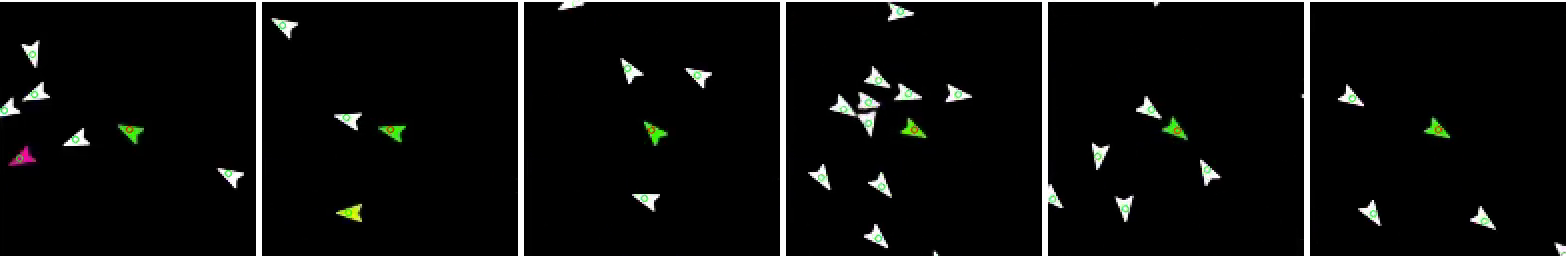} \\[3pt]
    \rotatebox[origin=c]{90}{\small Diff.\ forcing} &
      \includegraphics[width=0.92\linewidth]{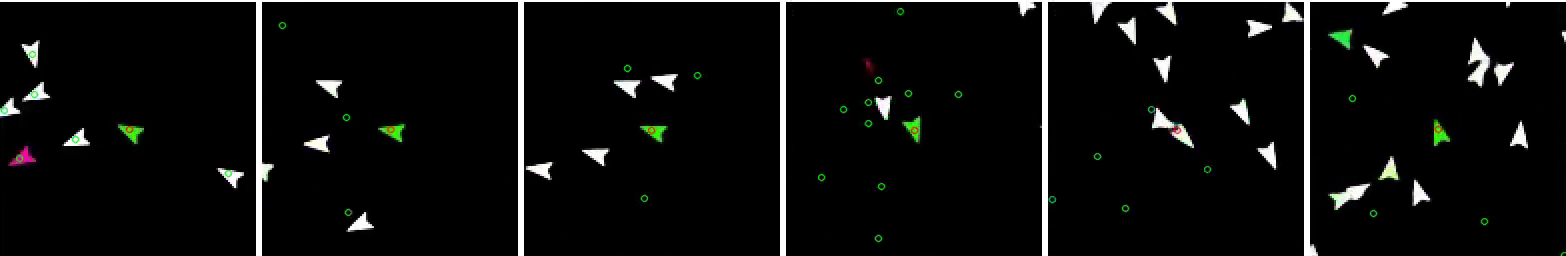} \\[3pt]
    \rotatebox[origin=c]{90}{\small Floor} &
      \includegraphics[width=0.92\linewidth]{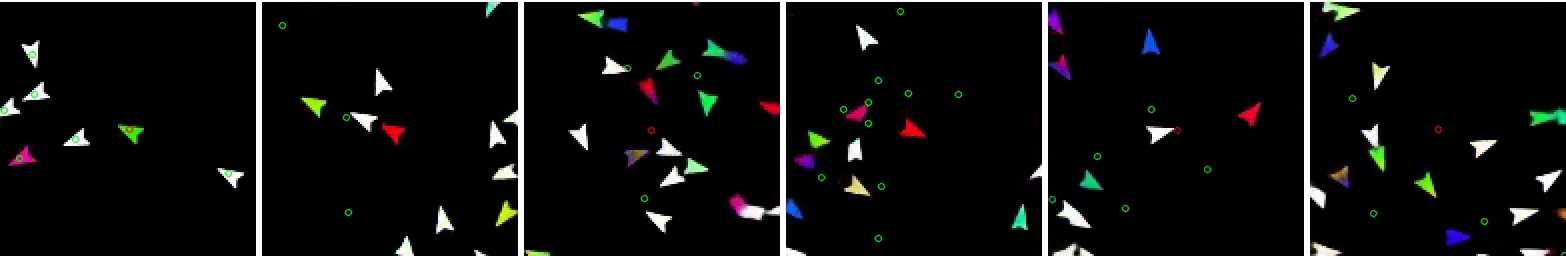} \\
  \end{tabular}
  \caption{One camera agent's egocentric view across a 10-second rollout, sampled
  approximately every $2$ s. The top row shows ground-truth latents decoded through
  the frozen autoencoder (ceiling), the middle row shows diffusion forcing, the model with
  the lowest cross-view displacement error, and the bottom row shows the floor, in which
  each view is rolled out independently by a single-agent model. All rows share the same
  ground-truth context, so the leftmost column is identical and all subsequent frames
  are generated. The floor hallucinates camera-agent colors and does not reliably
  preserve the focal agent's hue, whereas the diffusion-forcing model maintains the
  focal identity and renders most neighbors as uncolored boids.}
  \label{fig:filmstrip}
\end{figure}

Fig.~\ref{fig:filmstrip} shows one camera agent's egocentric view over a 10-second
rollout, sampled at approximately two-second intervals, for the VAE decode ceiling,
the diffusion-forcing model, and the independent single-agent floor. All three rows
begin from the same ground-truth context frames, so the first column is identical
across rows. All subsequent differences therefore come from the generated rollout.

The examples illustrate the distinction measured by our evaluation metrics. The
floor, which does not jointly model the camera agents, generates colored boids that
are not present in the episode and does not reliably preserve the focal agent's
identity. It therefore produces many independent sightings without corresponding
spatial agreement across views. The diffusion-forcing model instead preserves the
focal agent's identity and renders most neighboring boids as white, anonymous
agents. It makes fewer claims about the locations of other camera agents, but the
claims it does make are substantially more consistent across views. Both generated
models accumulate visual artifacts over the rollout, including softer boids and
small ring-like artifacts. These artifacts illustrate the accumulation of errors when the model relies entirely on its own generated frames as context.

\section{Reproducibility}
\label{app:repro}

\paragraph{Code and configuration.}
All experiments are controlled by YAML configuration files, one per experiment. The
baseline configuration reproduces Experiment 1, while each variant changes only the
flags corresponding to its additional mechanism. For tiled views these include
\texttt{tiled\_rope}, \texttt{tile\_grid=[2,5]}, \texttt{broadcast\_actions},
\texttt{shared\_timesteps}, and \texttt{use\_agent\_embed=false}. Diffusion forcing
uses \texttt{diffusion\_forcing}, and two-stage training uses \texttt{warm\_start}.

The Configuration column of Table~\ref{tab:experiments} names the configuration file
that defines each run. In the released checkpoints and evaluation outputs the same
seven runs appear under the identifiers \texttt{exp01\_baseline},
\texttt{exp02\_tiled}, \texttt{exp03\_diffusion\_forcing}, \texttt{exp04\_two\_stage},
\texttt{exp06\_tiled\_df}, \texttt{exp07\_tiled\_df\_two\_stage}, and
\texttt{exp08\_tiled\_two\_stage}, corresponding to Experiments 1 through 7 of
Table~\ref{tab:experiments} in that order. That numbering is not contiguous and does
not match the experiment numbers used in the paper, so the mapping given here is the
authoritative one.

\paragraph{Training schedule.}
Each run is bounded by wall clock rather than by a fixed epoch count. This
implements the fixed-budget protocol of Appendix~\ref{app:protocol}. Training is capped
at 2,860 minutes, and two-stage runs divide that cap into 710 minutes of single-agent
pretraining followed by 2,150 minutes of multi-agent training. Checkpoints are written
every 1000 steps so that a run stopped at the cap retains its final state. Stage 2
loads the single-agent weights through \texttt{train.warm\_start}, while
multi-agent-specific parameters are initialized from scratch and the optimizer is
reset. For runs combining diffusion forcing with two-stage training, diffusion forcing
is enabled during both stages so that the two stages share one objective.

\paragraph{Evaluation.}
Every checkpoint is evaluated with the same multi-view script under the protocol of
Appendix~\ref{app:eval}, which fixes the held-out episodes, the rollout length, the
number of sampling steps, and the ceiling and floor references. The floor reported
throughout the paper is the dedicated full-budget single-agent run, stored as
\texttt{single\_stream\_floor}. The released comparison tables also carry a column
named \texttt{floor} that holds a different and weaker model, namely the stage-1
pretrain of the two-stage run, and that column is not used anywhere in this paper.

\paragraph{Determinism.}
The simulator, data splits, and training procedure use fixed seeds, with seed 42
throughout. Episode seeds are derived from the base seed and episode index, making
the evaluation episodes and streamed training distribution reproducible from the
configuration files. Each visual condition uses one fixed autoencoder checkpoint, so
every run within a condition shares the same latent space.

\section{Compute Resources}
\label{app:compute}

All training and evaluation were performed on a SLURM cluster using one GPU per run,
either an NVIDIA A40 or A100, with 26 CPU cores and 90 GB of host memory. Every
experiment was allocated a single 48-hour array-job slot, so all runs received the same
wall-clock budget, and two-stage runs divide that slot between their two stages rather than
receiving additional time. In total, the experiments comprise fifteen runs, namely
the seven mechanism conditions of Table~\ref{tab:experiments}, the six agent-count
settings, the single-agent floor, and the gradient-background run, corresponding to
720 GPU-hours, or 30 GPU-days. That figure counts world-model training only, and the
autoencoder training and the evaluation passes described below are additional. Streaming data generation runs on the CPU alongside training, and data generation is limited by the simulator during the single-agent stage.

Evaluation is substantially cheaper than training. A complete evaluation pass over all
checkpoints takes approximately eleven
hours on a single GPU outside the cluster.

The video autoencoder was trained separately from the world-model campaign and on
different hardware, a single NVIDIA RTX 4090 rather than the cluster.

\section{Limitations}
\label{app:limitations}

The main limitations of this study are as follows.

\begin{itemize}
  \item \textbf{A single, deliberately simple environment.} The setting is a
    two-dimensional flocking simulation with deterministic dynamics for a fixed seed and
    relatively simple visual appearance. Conclusions about which mechanisms improve
    cross-view consistency may not transfer directly to visually rich,
    three-dimensional, or stochastic environments. The simplicity is intended to
    isolate the multi-agent consistency problem rather than to establish generality.

  \item \textbf{Single run per condition under a fixed budget.} Every model has
    approximately 27.5M parameters and a 48-hour single-GPU budget, and each
    condition is trained once. Therefore we cannot determine whether the observed differences would persist across repeated training runs.

  \item \textbf{Action-state consistency.} In our flocking environment, an agent's acceleration is determined by the state of its neighboring boids. We record these accelerations from the simulation and replay them during rollout. As FlockWorld generates a flock that diverges from the ground-truth trajectory, the neighboring boids around the focal agent may also differ from those in the original simulation. The replayed acceleration is therefore computed from a neighborhood that may differ from the one in the generated frames. This does not conflict with our goal of generating diverse, mutually consistent flock realizations, since we do not require the generated trajectory to match the ground truth. However, the replayed action may become inconsistent with the generated flock as the rollout progresses. Future work could address this limitation by closing the action-state loop and recomputing each agent's acceleration from the generated multi-agent state during rollout.

  \item \textbf{Known blind spots in the metrics.} All metrics pass through a
    hue-based boid detector, so the ceiling sits below perfect performance even for
    ground-truth decodes. Reciprocity depends on how many events occur, so it must be considered alongside the corresponding volume counts. Per-view fidelity captures both divergence from the reference trajectory, which is not necessarily undesirable, and genuine rendering errors.

  \item \textbf{The measurement ceiling depends on $P$, and not monotonically.} The measurement ceiling varies with the number of agents, $P$, for two different reasons. At large $P$, the agent palette uses hues spaced $1/P$ apart, making them harder to distinguish as the palette becomes more crowded. This affects even ground-truth decodes, as the reciprocity ceiling drops from $51.9$ at $P=10$ to $14.3$ at $P=20$ and $7.9$ at $P=30$ on the $0$ to $100$ scale of Table~\ref{tab:agentcount}. The displacement-error ceiling, which is not reported in Table~\ref{tab:agentcount}, also rises from $4.6$ pixels at $P=10$ to roughly $30$ pixels at $P=30$. 
  
  At the low end, however, hue crowding is not the problem. The reciprocity ceiling is also low at $P=2$, at $14.8$, even though the two hues are maximally separated. Instead, the limitation is co-visibility, since the two camera agents rarely appear in each other's crops. The $P=2$ reciprocity ceiling is based on only $212$ sightings across all 28 episodes, compared with $28{,}450$ at $P=10$.
  
  These results mean that identity-dependent metrics cannot be compared directly across different values of $P$. Each result must be interpreted relative to the measurement ceiling at the same $P$. This limitation comes from the hue-based measurement pipeline, not necessarily from the models. Comparing agent counts directly would require an identity cue that does not become harder to distinguish as $P$ increases and does not depend on ground-truth information.
  
  \item \textbf{The frozen autoencoder bounds everything.} World models cannot generate better reconstructions than the autoencoder itself, and the $16\times$ spatial compression limits the fine details of individual boids that can be represented.
\end{itemize}

\paragraph{Cost scaling.}
The training data itself is not the primary scaling constraint because clips are
generated on the fly. The effective training set is therefore not fixed, and the number
of unique clips is limited primarily by wall-clock training time rather than dataset
size.

The number of agents is the primary computational constraint. Each agent contributes
$S=64$ tokens per latent frame, and attention is unrestricted across agents within
each timestep, so a timestep's attention block spans $P\cdot S$ tokens. Under the
block-causal mask, frame $t$ attends to all frames up to and including $t$, so
summing over frames gives a total attention cost for a clip of $F$ latent frames of
\begin{equation}
\label{eq:attn-cost}
\sum_{t=1}^{F} (PS)(t\,PS) \;=\; (PS)^2\,\frac{F(F+1)}{2}
\;=\; \mathcal{O}\!\left(F^2P^2S^2\right).
\end{equation}
Increasing the number of agents from 10 to 30 raises the sequence length from 3,840
to 11,520 tokens and increases the per-timestep attention cost by roughly a factor of
nine, assuming the same model width and batch size. Thus, although the simulator
itself can efficiently support hundreds of boids, the dense cross-agent attention
used by FlockDiT becomes a major scaling bottleneck. Sparse attention over spatial
neighborhoods is a natural direction for future work because flocking interactions
are themselves local, but we do not evaluate such an approach here.

The gradient-background condition is covered by a single run and lacks a floor trained
under matching rendering, so its comparison is weaker than the black-background one, and
completing it requires training a condition-matched single-agent model. Future work
should also evaluate interactive rather than replayed-action generation, and scaling
past the $P=30$ we report will require relaxing the dense cross-agent attention of
Eq.~\eqref{eq:attn-cost}.

\section{Broader Impacts}
\label{app:impacts}

The main positive impact of this work is accessibility. FlockWorld provides a procedural environment for studying multi-agent world models without proprietary data or large-scale compute. The experiments use a 27.5M-parameter model and run on a single GPU for 48 hours. Together with the released code and configurations, this makes multi-agent modeling accessible to small research groups and undergraduate researchers. The simulator can also be varied in the number of camera agents, visual representation, and interaction radius.

Our evaluation also provides a general methodological contribution. It separates errors relative to a reference trajectory from disagreement between generated views, and reports consistency alongside rendered agent density. These controls help reveal cases where a model achieves a high consistency score for the wrong reason and can be applied beyond flocking.

The work also has a creative dimension, as it turns a simple artificial-life simulation into first-person visual experiences of collective motion, offering a potential tool for generative art.

The direct risks of our artifacts are limited. All data are procedurally generated, with no human or scraped data, and the models produce only $128\times128$ views of simple geometric boids. Scaling this approach to real-world settings could introduce additional concerns, such as privacy when modeling human crowds and increased computational costs as the number of agents grows. These concerns are outside the scope of this work.

\section{Existing Assets and Licenses}
\label{app:assets}

Three existing code assets are incorporated and adapted. The corresponding license
texts are included with the code release, and affected source files retain the
appropriate attribution in their header comments.

\paragraph{Solaris (Apache License 2.0).}
\url{https://github.com/solaris-wm/solaris}, commit \texttt{68e0ed3}. Our
flow-matching transformer is a PyTorch reimplementation of a model originally written
in JAX/Flax-nnx. We retain its factorized 3D RoPE split, RMS normalization,
sinusoidal timestep embedding, AdaLN-zero DiT block with six modulation parameters,
block-causal attention, patchify/unpatchify Conv3d scheme, and per-frame token
interleaving across agents. We remove the CLIP image-to-video cross-attention and
replace the mouse/keyboard action module with a two-dimensional steering acceleration
injected through AdaLN. Tiled RoPE, broadcast action conditioning, per-agent action
tags, and optional per-layer agent embeddings are additions introduced in our
implementation. Our flow-matching code reimplements the rectified-flow forward
process, velocity target, and $[0,1000]$ timestep range, and adds per-frame
diffusion-forcing timesteps and a context-preserving Euler sampler.

\paragraph{Wan 2.2 (Apache License 2.0).}
\url{https://github.com/Wan-Video/Wan2.2}, whose causal 3D video autoencoder follows the
architecture described in the Wan technical report~\cite{wanvae}. Copyright 2024-2025
Alibaba Wan Team Authors. The causal 3D video VAE is included with its copyright
header intact and reduced to the encoder, decoder, and wrapper used by this project.
The VAE checkpoints used in our experiments were trained from scratch on this
environment at a smaller configuration, using 8 latent channels with $16\times$
spatial and $4\times$ temporal compression. We do not load or redistribute released
Wan model weights.

\paragraph{cubeDhuang/boids (MIT License).}
\url{https://github.com/cubeDhuang/boids}. Copyright (c) 2023 Daniel Huang. The
flocking update loop, five-point boid geometry, and reference color mapping are
reimplemented in JAX as a batched, vectorized update. The reimplementation preserves
the vision radius, velocity-dot-product alignment bias, Reynolds-style steering with
maximum-force clamping, velocity drag, random heading noise, and minimum/maximum
speed limits.

\paragraph{MIRA (no public code release).}
MIRA~\cite{hu2026multiplayer} is reimplemented from its published description rather
than from source code, and is therefore credited by citation only.

\end{document}